\documentclass[10pt]{article}
\usepackage{spconf,amsmath,graphicx}

\usepackage{floatrow}
\newfloatcommand{capbtabbox}{table}[][\FBwidth]

\usepackage{blindtext}
\usepackage{balance}

\title{Language Model Bootstrapping Using Neural Machine Translation for Conversational Speech Recognition}
\name{\textit{Surabhi Punjabi, Harish Arsikere, Sri Garimella}}
\address{Alexa Machine Learning, Amazon, Bangalore, India \\
\{spunjabi, arsikere, srigar\}@amazon.com}

\begin{document}
\ninept
\frenchspacing
\maketitle
\begin{abstract}
Building conversational speech recognition systems for new languages is constrained by the availability of utterances capturing user-device interactions. Data collection is expensive and limited by speed of manual transcription. In order to address this, we advocate the use of neural machine translation as a data augmentation technique for bootstrapping language models. Machine translation (MT) offers a systematic way of incorporating collections from mature, resource-rich conversational systems that may be available for a different language. However, ingesting raw translations from a general purpose MT system may not be effective owing to the presence of named entities, intra sentential code-switching and the domain mismatch between the conversational data being translated and the parallel text used for MT training. To circumvent this, we explore following domain adaptation techniques: (a) sentence embedding based data selection for MT training, (b) model finetuning, and (c) rescoring and filtering translated hypotheses. Using Hindi language as the experimental testbed, we supplement transcribed collections with translated US English utterances. We observe a relative word error rate reduction of 7.8-15.6\%, depending on the bootstrapping phase. Fine grained analysis reveals that translation particularly aids the interaction scenarios underrepresented in the transcribed data.
\end{abstract}
\section{Introduction}
\label{sec:intro}
Bootstrapping an automatic speech recognition (ASR) system for a new language involves significant data collection and transcription overhead. For factored ASR systems, where the acoustic model (AM) and language model (LM) are trained independently, the LM can be trained with additional text-only corpora to boost performance. This is especially helpful during the initial stages of model development. For a new language the typical supplemental LM sources include Wikipedia, news portals, blogs etc., which can be incorporated along with the limited transcribed data to circumvent the issue of cold start. However, for conversational agents like Alexa, Siri the utterances are usually short, goal directed and contain several named entities like song title, artist name etc., e.g. \textit{Play Moonlight Sonata by Beethoven}. This informal interaction style, characteristic of conversational data, is absent in online text sources, thereby rendering them less effective for this task. As a result, LM building relies mostly on transcribed data and its performance is restricted by the speed of manual transcription and annotation.

There has been a growing interest in the area of data augmentation for ASR language modeling. Previous studies include training a recurrent neural network (RNN) based LM on transcriptions and using it to generate synthetic samples for augmentation \cite{GorinIS2016}. SeqGAN, a generative adversarial model for sequences, has been employed for pretraining a code-switched LM \cite{SGargIS2018}. However, a precondition for the successful generalization of these neural generative models is the availability of a substantial amount of in-domain utterance text for training, which itself is the bottleneck during the bootstrapping phase. 
  
Utterances from mature conversational systems, for example in English provide a rich source of information. They are both \textit{in-domain}, since they capture actual user interaction patterns of varying complexity, and \textit{large-scale},  owing to prolonged usage. Translation offers an elegant and cost-effective solution for leveraging this existing data. Devising techniques for systematically incorporating translated data can be instrumental for achieving the rapid language expansion goal for ASR, by alleviating the prohibitively high requirements for data collection during bootstrapping. 

The area of machine translation has witnessed sustained research efforts \cite{CHO2014,Bahdanau2015JOINTLY,GNMT}. It is also amongst the first success stories of the end-to-end neural paradigm for sequence modeling. Conventional phrase based statistical machine translation (SMT) \cite{PBSMT} has shown to be outperformed by attention based recurrent encoder-decoder models \cite{GNMT} and transformer networks comprising self-attention and feed forward network blocks \cite{TRANSFORMER}. 

Data augmentation via SMT has been explored in the past for keyword spotting \cite{GorinIS2016} and ASR \cite{ICELANDIC, ENGMAND}. These studies primarily focus on incorporating raw translation output as a component in the LM. However, in our initial experiments we observed that directly ingesting translations generated from off-the-shelf MT models results in a suboptimal performance for conversational data. This could be attributed, in part, to the domain mismatch between the MT training data comprising parallel text from web sources and the informal style interaction data used for translation. This observation of MT output being sensitive to the mismatch in training and inference data distributions is consistent with previous studies on MT adaptation \cite{ADAPTATIONSURVEY}. 

Statistical post-editing for improving the quality of SMT outputs has been investigated \cite{ROM}. In a recent work on bootstrapping natural language understanding systems using translations \cite{MTNLP}, SMT is employed for generating initial translations, followed by the use of source-target alignments to retain and resample named entities. These post-editing approaches can minimize the undesired named entities conversions for SMT, yet the bigger issue of domain mismatch still remains open. 

In this work, we explore the synergies between neural machine translation and speech recognition for data augmentation. We work towards bootstrapping Hindi ASR system. Along with the limited availability of representative transcribed data, an additional challenge in this setting is that of code-switching. In typical Hindi utterances people often code mix with English within a sentence. The techniques explored in this work are however generic, and Hindi is chosen as a testbed for its complexity.

We evaluate different architectures for building English to Hindi (EN$\rightarrow$HI) translation models and elaborate on the pitfalls associated with using off-the-shelf translations. Some initial gains are observed by inferring alignments from attention weights, an approach that enables preserving and resampling the named entities. This technique is further extended to simulate code-switching in the translated data.

We then delve into the deeper issue of domain inconsistency. To this end, we develop a data selection strategy for MT model training based on in-domain similarity. This is an extension of \cite{WANG2017a} for the fully unsupervised setting. We also assess model finetuning approach by adding parallel in-domain synthetic pairs. For further adaptation, a statistical LM built using transcribed data is used for rescoring the decoded translation beams. Finally, different quality metrics are compared for retaining only the high quality translations in the final translation component.

A comparative evaluation of the translation-augmented LM is performed against baselines built from only transcribed data at various stages of bootstrapping. To the best of our knowledge, this work is the first investigation of the efficacy and challenges associated with neural machine translation for conversational speech recognition.

\section{Machine translation for Data Augmentation }

\label{sec:Method}
\subsection{Building translation model}
\label{ssec:transsystem}
Neural machine translation (NMT) is the dominant paradigm for current MT research. We assess the popular neural architectures for the task of building an EN$\rightarrow$HI translation model. Sequence-to-sequence framework with attention, proposed in \cite{Bahdanau2015JOINTLY}, comprises two recurrent neural networks: \textit{Encoder}, which reads the source sentence tokens $(x_{1}, x_{2},..x_{t})$ to generate continuous representations $(h_{1}, h_{2},..h_{t})$, and \textit{Decoder}, which outputs symbols $(y_{1},y_{2}...y_{n})$, conditioned on the previous outputs as well as a context vector \textbf{c}, derived as the weighted sum of the encoder hidden states \textbf{h}. 

Transformer networks proposed in \cite{TRANSFORMER} eliminate recurrence in favor of parallelism and rely solely on attention. Here the encoder and decoder comprise stacked self-attention and fully connected layers. These networks represent the current state-of-the-art for NMT.

\begin{figure}[!hbtp]
\begin{center}
\includegraphics[scale=0.248]{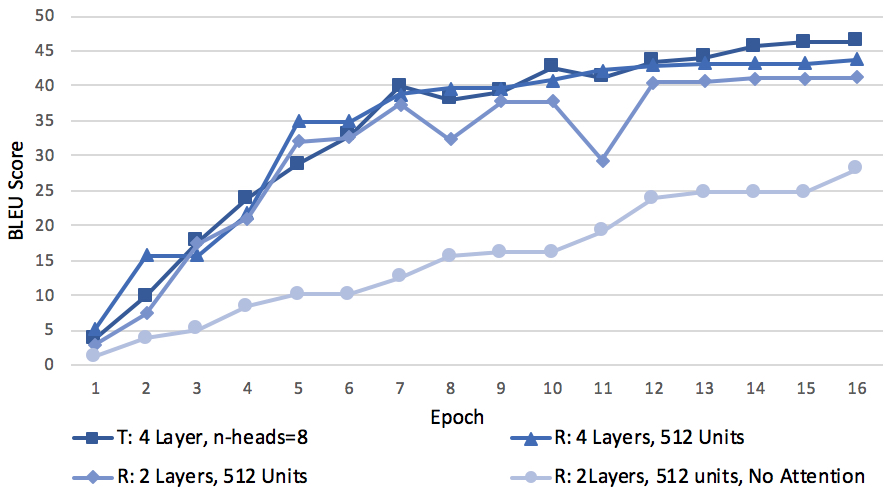}
\end{center}
\vspace*{-8mm}
\caption{ \textit{BLEU score over epochs. Transformer and recurrent architectures are represented by labels T and R respectively.}\label{NMTTrain}}
\end{figure}

For training the translation models, we use a corpus of 8.4M parallel (EN, HI) sentence pairs prepared by crawling different web sources. We employ BLEU (Bilingual Evaluation Understudy) \cite{BLEU} score to assess the translation quality. Fig. \ref{NMTTrain} captures the performance of these models for different configurations. The best performing recurrent encoder-decoder with attention model achieves a BLEU score of 43.8 as compared to 46.4 achieved by the transformer architecture. 

\begin{figure*}[!hbtp]
\begin{center}
\includegraphics[scale=0.195]{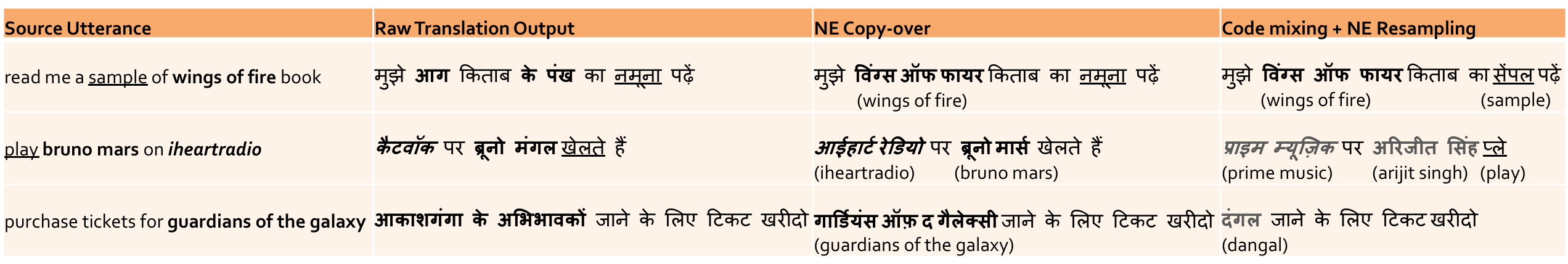}
\end{center}
\vspace*{-5mm}
\caption{ \textit{Examples of raw translations and postprocessed outputs. NE is a shorthand for named entities. Aligned source and target tokens are indicated by similar highlighting. Underlined tokens indicate English tokens in the simulated code mixing. }\label{anec}}
\end{figure*}
\vspace*{-5mm}
\subsection{Choice of NMT architecture}
Postprocessing translations generated by an MT model can be facilitated by the information of source-target alignments. In case of SMT this is straightforward owing to the fact that the alignment model is learnt explicitly. In case of NMT models, the separate components of the conventional SMT are folded into an all neural architecture. With this simplification however, assessing which source token is responsible for generating a target token becomes tricky.

Attention mechanism in the recurrent encoder-decoder architecture serves as an implicit alignment model, allowing the decoder to focus on relevant source segments. Eq. \eqref{eq:1} captures decoder state $s_{i}$ as a function of previous state $s_{i-1}$, previous output $y_{i-1}$ and the context vector $c_{i}$. Here $T_{x}$ represents the number of tokens in the source sequence, and $f$ denotes some nonlinear function. Notice that the attention weight $\alpha_{i,j}$ determines the weight assigned by decoder at time step $i$ to encoder hidden state $h_{j}$. 
\begin{equation}
\label{eq:1}
s_{i} = f(s_{i-1}, y_{i-1}, c_{i}) \text{ where, } c_{i} = \sum_{j=1}^{T_{x}} \alpha_{ij}h_{j}
\end{equation} 
Deriving alignments is known to more challenging for transformer networks with self-attention and multiple attention heads. There has been some recent work for alleviating this issue by explicitly adding an alignment head to the base architecture \cite{TRANSALI2018}.

Owing to the relative ease of alignment extraction, we make the modeling design decision of using the recurrent encoder-decoder networks with attention for our NMT experiments.

\vspace*{-3mm}
\subsection{Incorporating raw translations: An initial study}
As an initial experiment, we translated user interaction sentences from US English collections to Hindi and directly  ingested the raw translations for LM training. However, this strategy resulted in a very high perplexity LM. Upon further analysis, we found three key explanatory factors. First, typical user interactions with voice controlled agents, e.g. song requests, contain several named entities. The general purpose EN$\rightarrow$HI MT system generates translations for those entities as well, which is not desirable. The second factor is the absence of code-switching in translations, which are purely in Hindi, owing to the nature of the training data. Given the extent of intra-sentential code mixing in conversational Hindi, it seems imperative for the translations to capture it as well, in order to add value to the downstream language modeling task. Finally, the most challenging nuance is the out-of-domain nature of MT training data (news items, wiki articles, etc.), which results in a lack of informal interaction style in the generated translations, an inherent attribute of user-device interactions. This domain mismatch issue has been observed in other machine translation settings as well. 

\vspace*{-3mm}
\subsection{Post-editing translations}
We use the attention weights derived while decoding to approximate alignments. Each target token generated by the decoder is considered to be \textit{aligned} with the source token corresponding to the encoder position with maximum attention weight. Metadata in the source text annotations, e.g. song name, artist name etc., is used to identify named entities (NE). Using these alignments and annotations, the following post-editing steps are performed. 1) \textit{NE copy-over}: source tokens corresponding to named entities are simply retained as-is in the output, 2) \textit{NE resampling}: named entities are resampled with local Hindi catalogs, since the trending entities vary by geography and, 3) \textit{Code mixing}: code-switching is simulated in the translations, by probabilistically copying over source English tokens. The probability of retaining a token is set to be directly proportional to the smoothed relative frequency in the transcribed Hindi collections. Fig. \ref{anec} provides examples of raw translations and the outputs of the post-editing step.
\begin{figure*}[!hbtp]
\begin{center}
\includegraphics[scale=0.08]{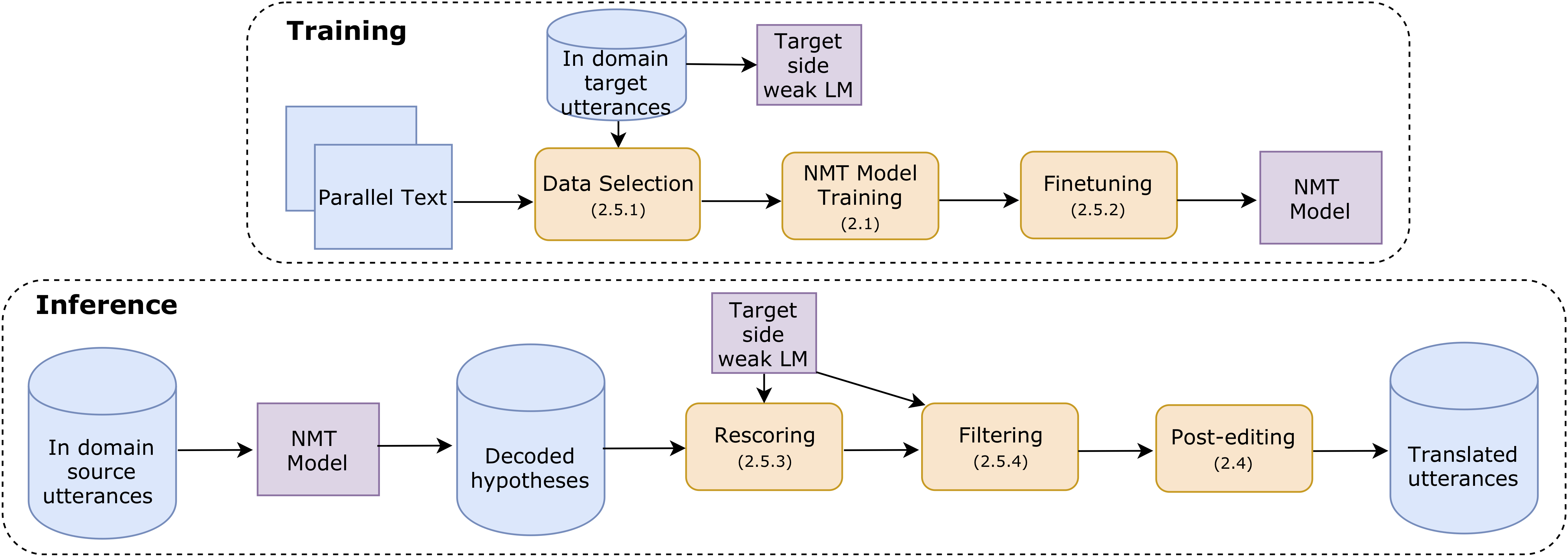}
\end{center}
\vspace*{-5mm}
\caption{ \textit{Overview of the model training, adptation and postprocessing pipeline. $(.)$ indicates the corresponding section describing the approach.}\label{schematic}}
\end{figure*} 
\subsection{Domain adaptation}
Domain mismatch between the training and inference data presents a more subtle challenge with the translations not reflecting colloquial usage. Most of the prominent approaches for MT model adaptation, like backtranslation \cite{SENNRICH}, shallow fusion \cite{FUSION} etc., assume presence of a large target side monolingual corpus for boosting the fluency of translations. Such an assumption cannot be made during model bootstrapping, where only limited target in-domain data (transcribed collection) is available for adaptation. With this constraint in place, we experiment with four broad classes of techniques for adaptation: 1) data selection for MT training, 2) model finetuning, 3) rescoring, and 4) filtering translations.
\vspace*{-4mm}
\subsubsection{Data selection for MT training}
Selecting NMT training samples similar to the in-domain data from the out-of-domain parallel corpus has been explored in \cite{WANG2017a}. The central idea of this work is to use both in-domain and out-of-domain parallel corpora to train an NMT system. Similarity between the learnt encoder representations of these sentences is used to define their semantic closeness. Sentence selection is based on the relative similarity to the in-domain versus out-of-domain vector centers. 

We adopt a similar data centric approach for adaptation. In our setting however, in-domain parallel corpus is not available. Hence, representing sentences via NMT encoder embeddings is not feasible. In order to facilitate data selection, we resort to the approach of learning unsupervised sentence embeddings for quantifying the closeness of  target side MT training sentences with the in-domain transcriptions available.

We compare the following techniques for generating unsupervised sentence representations: (1) unweighted averaging of word vectors, (2) smooth inverse frequency \cite{SIF}, where the sentence embeddings are computed as a weighted average of word vectors followed by removal of the projection on the first singular vector, and (3) language agnostic sentence representations (LASER) \cite{LASER}, which is an open source pre-trained biLSTM encoder for generating multilingual sentence embeddings that generalize across languages and NLP tasks.
The first approach is appealing owing to its simplicity. In the second technique, taking word frequency into account is the distinguishing factor. Potential cross-lingual generalization is the advantage offered by the third approach. We use  FastText \cite{FASTTEXT} for learning word vector representations.

Using each of these approaches, we generate sentence embedding vectors for in-domain ($F_{in}$) and out-of-domain ($F_{out}$) target side sentences. Along similar lines as \cite{WANG2017a}, data selection is based on the relative distance $\delta_{s}$ of a sentence vector $v_{s}$ w.r.t. the in-domain and out-of-domain centroids $C_{F_{in}}$ and $C_{F_{out}}$ respectively, indicated by Eq. (\ref{eq:2}). A lower value of $\delta_{s}$ implies higher resemblance with in-domain data. 
\vspace*{-2mm}
\begin{equation}
\label{eq:2}
\delta_{s} = d(v_{s}, C_{F_{in}}) - d(v_{s}, C_{F_{out}}) 
\end{equation}
\begin{equation*}
 \text{where, } C_{F_{in}} =\frac{\sum_{f \in F_{in}} v_{f}}{|F_{in}|} \text{,  } C_{F_{out}} =\frac{\sum_{f \in F_{out}} v_{f}}{|F_{out}|}
\end{equation*} 
\begin{equation*}
 \text{and, } d(x, y) = || x - y ||_{2} 
\end{equation*} 
\vspace*{-8mm}
\subsubsection{Model finetuning}
Backtranslation \cite{SENNRICH} is a popular approach for adaptation where a target-to-source translation model is learnt on the parallel corpus and is used to translate the unpaired target-side monolingual data. The resulting synthetic (source, target) pairs are leveraged for the original source-to-target model training. In the absence of a large target monolingual corpus, we resort to an alternate approach for synthetic corpus generation. We generate pseudo pairs by translating a portion of US English utterances using an initially trained NMT model, and perform post-editing to retain named entities. With this additional parallel data, the model is further trained for a certain number of epochs. As we discuss in Section \ref{sec:Result}, this type of finetuning is susceptible to overfitting.

\subsubsection{Rescoring with in-domain LM}
In order to further boost the fluency of translations, the hypotheses obtained after beam search decoding are rescored using an n-gram LM built from the in-domain transcribed data. The score of a translation hypothesis is computed as a weighted sum of the MT decoding score and the LM score. The choice of n-gram LM for rescoring is motivated by its robustness under low-resource conditions as compared to RNNLM.

\subsubsection{Filtering translations}
As a final step, we attempt to remove the spurious translations by performing filtering based on a quality measure. The main challenge here is to define the ``goodness" of a translated output. A potential candidate is the approach of using the score assigned by the MT model. That is, the product of conditional probabilities of the output tokens generated by the MT decoder can be used as a proxy to define well formed translations. We also consider the approach of using the statistical LM built using transcribed data to assess the quality of translation data. Using each of these scores, we retain the top-x percentile of the translation output. 

An overview of the adaptation and postprocessing pipeline is provided in Fig. \ref{schematic}, with section numbers indicated against each component. 

\section{Results and Discussion}
\label{sec:ResultAndDiscuss}

\subsection{Experimental setup}
We conduct experimental evaluation using upto 180 hours of Hindi-English code-switched speech for training. This dataset comprises 200K Hindi utterances collected using Cleo, an interactive skill that enables users to teach local languages to voice assistants via prompts. These prompts cover use cases like song requests or knowledge related questions. These natural utterances represent the transcribed in-domain component in our experiments. We follow the factored ASR architecture, and the AM is a hybrid DNN-HMM model, trained on log filter bank energy (LFBE) features extracted at 10 ms intervals for a 25 ms analysis window. LM is built by learning a 4-gram model with Katz smoothing on the training data.

The baseline LM is built using the in-domain transcriptions only. Translation component is procured by translating 9.8M US English utterance transcripts using the trained NMT models followed by adaptation and post-editing. For the evaluation candidates, the LM is built by a linear interpolation of the transcribed and translated components. The interpolation weights are tuned to minimize the perplexity of a held-out in-domain dataset. We assign a floor interpolation weight of 0.25 to the translation component to ensure that it receives sufficient representation. For all the MT experiments except rescoring, we employ greedy decoding, i.e. beam size of one. The test set comprises 37K utterances. We report relative word error rate reduction (WERR) w.r.t. models built using baseline LM.

\subsection{Results}
\label{sec:Result}

\begin{table*}[!ht]

\bigskip
\centering
\begin{tabular} {c|c|c|c}

\hline
\textbf{Postprocessing} & \textbf{Approach} &  \textbf{PPL} & \textbf{Relative WERR \%} \\
\hline
\hline
None & Raw translations  & 11941.08 & -1.81		 \\
\hline
Post-editing  & NE copy-over &  2889.45	 & 2.36	 		\\
&  NE resampling &  1241.52 & 4.62  		\\
&  Code mixing + NE resampling &  936.64 &  5.83		\\
\hline

\end{tabular}
\\
\vspace*{-5mm}

\caption{\label{table:post-edit} \textit{ Relative WERRs (\%) with different post-editing techniques. Perplexity (PPL) is evaluated on a held-out in-domain dataset. Relative WERR captures the WER reduction w.r.t baseline trained on transcribed data only.}}
\normalsize
\end{table*}
\begin{table*}[!ht]
\centering

\begin{tabular} {c|c|c|c}

\hline

\textbf{Adaptation} & \textbf{Approach} &  \textbf{PPL} & \textbf{Relative WERR \%} \\
\hline
\hline
Data selection &  Unweighted avg. (BLEU: 29.1) & 662.33 & 6.94			\\
 (BLEU original model: 43.8) &   SIF (BLEU: 37.8)	 & 686.97	 & 7.23			\\
&   LASER (BLEU: 37.4)	 & 704.12	 & 7.14		\\
\hline

Rescoring & beam-size=5  & 792.92 & 6.28	 \\
& beam-size=20 & 852.16 & 5.88	 \\
\hline

Model finetuning & n-epochs=3 & 726.62 & 6.84	\\
& n-epochs=10 & 983.64 & 5.23	\\
\hline

Filtering translations & MT score - top 85\% & 1109.44 & 4.82 \\
 & MT score -  top 75\% & 1327.56 & 3.37 \\
  & MT score -  top 65\% & 1426.18 & 2.16 \\
\cline{2-4}
  & SLM score - top 85\% & 793.73 & 6.33 \\
 &SLM score - top 75\% & 892.92 &  6.82	\\
  &SLM score - top 65\% & 878.16 &  5.94\\
\hline

Combined & (i) SIF selection + Rescoring + SLM score - top 75\% &  584.24 & 7.62 \\
 & (i) + Model finetuning &  564.06  & 7.86 \\
\hline

\end{tabular}
\\
\vspace*{-5mm}

\caption{\label{table:mainResults} \textit{ Relative WERRs (\%) with different NMT adaptation strategies. Note that these results include the effect of NE resampling and code mixing techniques.}}
\normalsize
\end{table*}

Table \ref{table:post-edit} captures the performance of post-editing techniques. The approach of ingesting raw translations without any postprocessing yields a high perplexity translation component, resulting in negative WERR. We observe consistent improvements by introducing attention weight based post-editing. \textit{NE copy-over} alone reduces the perplexity significantly. This, coupled with \textit{NE resampling} and \textit{code mixing}, results in a 5.83\% WERR.

In Table \ref{table:mainResults}, we assess the impact of each adaptation approach followed by post-editing. For \textit{MT training data selection}, we retain only top-25\% (out of 8.4M) sentences w.r.t. their relative similarity with in-domain data. This reduction in training data impacts the BLEU score adversely. Amongst the sentence representation techniques, LASER and SIF embeddings outperform the unweighted averaging approach in terms of BLEU score. Interestingly, while the unweighted averaging achieves lowest perplexity on held-out in-domain dataset, the gains don't carry over while measuring overall ASR performance. SIF embedding based selection achieves the highest WERR of 7.23\%, followed closely by LASER encoder representation. 

\textit{Rescoring} the decoded beams using transcription based LM yields a WERR of 6.28\% for a beam size of 5. In these experiments, a relative weight of 0.3 is assigned to LM for overall score computation. Increasing the beam size from 5 to 20 leads to a drop in WERR, suggesting that during decoding, the head portion of the translation output contains hypotheses helpful for improving naturalness, and increasing beam width can result in higher confusability. 

For the model \textit{finetuning} approach, the number of additional training epochs is an important parameter. We observe a WERR of 6.84\% when this parameter is set to 3, as compared to 5.23\% for 10 epochs. Increasing the number of passes on the synthetic data generated using an initially trained model perpetuates the effect of model reinforcing its own errors. This potential overfitting makes early stopping imperative.

In the experiments focusing on \textit{translation output filtering}, MT score did not turn out to be an effective metric for quality evaluation, indicated both by perplexity and WERR. We obtain interesting insights by ranking translations using in-domain LM scores. A WERR of 6.82\% is observed by retaining only top-75\% translations. Making filtering conservative beyond this point degrades performance. One caveat of the LM guided filtering approach is that the patterns which are underrepresented in the initial collections receive low LM scores. This could explain the drop in WERR when moving to top-65\% translations; since the transcribed volume used for LM training is itself small, some of the discarded patterns could have been complementary for the overall ASR performance. 

Combining the SIF selection, finetuning, rescoring and LM based filtering approach results in a relative WERR of 7.86\%.

\subsection{Impact on different interaction scenarios}
\label{sec:domain}

Cleo prompts cover multiple interaction use cases. In order to derive fine grained insights into the effect of translations,  we study WERR on test utterances manually categorized into scenarios. Nearly 70\% of the test utterances fall into one of the nine interaction scenarios mentioned in Table \ref{table:scenarios}. In order to isolate the gains 
obtained from post-editing and adaptation, we study both the post-editing and combined WERR. We also analyse the proportion of named entities for utterances belonging to each of these scenarios. Following observations can be made from this analysis.

\begin{table*}[!ht]

\bigskip
\centering
\begin{tabular} {c|c|c|c|c|c}

\hline
\textbf{Coverage} & \textbf{Interaction} & \textbf{Post-editing} &  \textbf{Combined } &  \textbf{Adaptation  }   & \textbf{NE\%} \\
\textbf{(In transcribed collections)} & \textbf{scenario} & \textbf{WERR} \% & \textbf{WERR} \% & \textbf{contribution \%} & \\
\hline
\hline
Low & Books & 	5.75  & 	7.26  &	20.80  &	34.74  \\ 
 &  Communication	& 3.82 &	5.98 & 	36.12	& 11.19	\\	
&  Weather &	3.23	& 6.85	& 52.84	& 7.63	\\
&  Shopping	 & 7.86	 & 10.84	 & 27.49  & 	52.94   \\
\hline
Moderate &  Knowledge &	6.36 & 	9.54	& 33.34	& 31.60 \\
 & Video	& 6.44	&  8.52 & 	24.41	&  39.36	 \\
& Home Automation	&  5.68 & 	7.94	&  22.63 & 	5.81 \\
\hline
High & Notifications  &	4.65	 & 7.06  &	34.14   & 5.66	   \\
& Music &	5.74 &	7.48	 & 23.26	& 47.62  \\

\hline
\end{tabular}
\\
\vspace*{-4mm}

\caption{\label{table:scenarios} \textit{Relative WERR \% by interaction scenarios captured along with the extent of coverage in transcribed collections. Named entity proportion in the utterances is given by NE \%. Adaptation contribution \% captures the relative contribution of adaptation towards WERR. For e.g., with a 5.75\% post-editing WERR, adaptation yields an additional 1.51\% WERR towards a combined WERR of 7.26\%, i.e. 20.80\%.}}
\normalsize
\end{table*}

\begin{itemize}
	 \item Higher WERR is observed for scenarios such as shopping and knowledge related queries, which are not well represented in the transcribed collections as compared to their popular counterparts like song requests and notifications, suggesting that translations can effectively complement the transcribed data.
	 \item WERR shows positive correlation with the percentage of named entities (Pearson correlation: 0.647, p-value: 0.059). This observation is consistent with the results in the previous section where postprocessing alone demonstrates a significant WERR.
	 \item The relative contribution of adaptation towards WERR is higher for scenarios with smaller named entity footprint, e.g. weather. Hence, the gains from postprocessing and adaptation seem to be complementary across conversation scenarios.
 \end{itemize}

\subsection{Impact of floor weight for interpolation}
\label{sec:Floor weight sweep}
The final augmented LM is an interpolated n-gram model, with the probability of an n-gram computed as a weighted sum of probabilities assigned by transcribed and translation components. Since the tuning data for determining interpolation weights comprises transcribed utterances only, the translation component may receive a low weight owing to domain mismatch.

The purpose of this investigation is to observe the effect of changing the floor weight parameter for the translation component, which provides a lever to override its relative importance in the interpolated LM. As seen from Table  \ref{table:FLR}, the overall PPL increases as we increase the floor weight. However, WERR demonstrates fluctuation with varying floor weights: a low weight renders the translation component ineffective whereas a high value undermines the transcription component. Floor weight sweep can provide empirical guidance for adjusting this parameter.

\begin{table}[!ht]
 
\begin{tabular} {|c|c|c|}

\hline
\textbf{Floor} & \textbf{Interpolated} & \textbf{WERR $\%$} \\
\textbf{weight} & \textbf{PPL} & \\
\hline

0.1  &  50.28 & 5.78   \\
0.15 & 51.24 & 7.04 \\
0.25 & 52.36 & 7.86 \\
0.3 & 53.37 & 7.49 \\
0.4 & 56.34 & 6.58 \\
\hline
\end{tabular}

\vspace*{-4mm}
 \caption{\label{table:FLR} \textit{PPL and relative WERRs ($\%$) with varying floor interpolation weights for the translation component in the 180 hour setup.}}
\normalsize
\end{table}

\vspace*{-5mm}
\subsection{Impact of in-domain data volume}
\label{sec:Data sweep}
We now attempt to address the following question: what are the relative gains provided by the translation data during different phases of bootstrapping? In particular, we measure the WERR between the baseline and translation-augmented LMs, by varying the in-domain transcribed utterances from 10K to 200K. We observe that the combined WERR after post-editing and adaptation increases from 7.86\% to 15.65\% as the amount of in-domain data reduces. Note that in this experiment, we use the same AM trained on 180 hours data, in order to precisely study the effect of data augmentation for LM. The WERR we report is hence an underestimate, and will probably be much higher, if the AM was trained using similar levels of transcribed  data. These findings, summarized in Table \ref{table:DATA}, suggest that the neural MT supplements can especially aid initial stages of model development.

\begin{table}[!ht]

\begin{tabular} {|c|c|}
\hline
\textbf{Transcribed } & \textbf{WERR $\%$} \\
\textbf{Volume} & \\
\hline

10K & 15.65 \\
20K & 13.18 \\
50K & 9.42 \\
100K & 8.98 \\
200K & 7.86\\
\hline

\end{tabular}  
\vspace*{-4mm}

\caption{\label{table:DATA}   \textit{  Relative WERRs ($\%$) with varying levels of in-domain transcribed data.}}
\normalsize
\end{table}

\vspace*{-6mm}

\section{Conclusion}
\label{sec:conclusion}
In this work, we explored the key challenges associated with using NMT for LM data augmentation in a conversational, code-switched setting. Using a combination of post-editing and domain adaptation techniques, we demonstrated a relative WERR of 7.8\% for 180 hours of transcribed data. We examined the performance trajectory along different bootstrapping phases, and observed relative WERR of upto 15.6\% with reduced transcription volumes. A further drilldown of WERR by interaction scenarios provided interesting insights into the gains derived from translation as a function of proportion of named entities and relative representation in the transcribed data. This experimental evidence establishes the efficacy of using translations for supplementing transcribed collections in the early stages of model development, a strategy which could be instrumental for rapid language expansion. Though Hindi is used as an experimental testbed in this work, the techniques presented are generic and can be leveraged for bootstrapping other languages as well. Exploring semi-supervised and unsupervised translation is a promising future direction, especially for the low resource languages.
\balance


\bibliographystyle{IEEEbib}
\bibliography{strings,refs}

\end{document}